\documentclass[letterpaper, 10 pt, conference]{ieeeconf}

\IEEEoverridecommandlockouts    

\makeatletter
\let\NAT@parse\undefined
\makeatother

\usepackage{cite}
\usepackage{amsmath,amssymb,amsfonts}
\usepackage{bm}
\usepackage{algorithmic}
\usepackage{graphicx}
\usepackage{pifont}
\usepackage{textcomp}
\usepackage{hyperref}
\usepackage{multirow}
\usepackage{balance}
\usepackage[table,xcdraw]{xcolor}
\usepackage{url}
\usepackage{comment}
\usepackage{color}
\usepackage{latexsym}
\usepackage{mathtools}
\usepackage{caption}
\usepackage{subcaption}
\usepackage{orcidlink}

\title{\LARGE \bf
Sim2Real Diffusion: Leveraging Foundation Vision\\Language Models for Adaptive Automated Driving
}

\author{Chinmay Samak$^{\star \dagger}$ \orcidlink{0000-0002-6455-6716}, Tanmay Samak$^{\star \dagger}$ \orcidlink{0000-0002-9717-0764}, Bing Li$^{\dagger}$ \orcidlink{0000-0003-4987-6129}, and Venkat Krovi$^{\dagger}$ \orcidlink{0000-0003-2539-896X}
\thanks{$^{\star}$These authors contributed equally.}
\thanks{$^{\dagger}$Department of Automotive Engineering, Clemson University International Center for Automotive Research (CU-ICAR), Greenville, SC 29607, USA.
{\tt\small {\{\href{mailto:csamak@clemson.edu}{csamak}, \href{mailto:tsamak@clemson.edu}{tsamak}, \href{mailto:bli4@clemson.edu}{bli4}, \href{mailto:vkrovi@clemson.edu}{vkrovi}\}@clemson.edu}}}%
}

\begin{document}

\maketitle
\thispagestyle{empty}
\pagestyle{empty}


\begin{abstract}
Simulation-based design, optimization, and validation of autonomous vehicles have proven to be crucial for their improvement over the years. Nevertheless, the ultimate measure of effectiveness is their successful transition from simulation to reality (sim2real). However, existing sim2real transfer methods struggle to address the autonomy-oriented requirements of balancing: (i) conditioned domain adaptation, (ii) robust performance with limited examples, (iii) modularity in handling multiple domain representations, and (iv) real-time performance. To alleviate these pain points, we present a unified framework for learning cross-domain adaptive representations through conditional latent diffusion for sim2real transferable automated driving. Our framework offers options to leverage: (i) alternate foundation models, (ii) a few-shot fine-tuning pipeline, and (iii) textual as well as image prompts for mapping across given source and target domains. It is also capable of generating diverse high-quality samples when diffusing across parameter spaces such as times of day, weather conditions, seasons, and operational design domains. We systematically analyze the presented framework and report our findings in terms of performance benchmarks and ablation studies. Additionally, we demonstrate its serviceability for autonomous driving using behavioral cloning case studies. Our experiments indicate that the proposed framework is capable of bridging the perceptual sim2real gap by over 40\%.\\%
\end{abstract}



\section{Introduction}
\label{Section: Introduction}

Modern automated driving systems (ADS), especially vision-based algorithms, are increasingly data-driven, relying on large volumes of diverse, high-quality data for training and validation. However, real-world data collection is expensive, time-consuming, and often unsafe or infeasible -- particularly for rare, safety-critical scenarios. Simulation frameworks address these limitations by enabling safe, scalable, controllable, and cost-effective generation of synthetic datasets across varied conditions, which is essential for robust and safe validation \cite{CornerCaseAnalysis2021}. Simulations also ensure reproducibility and standardized benchmarking \cite{9340902}, while their parallelization capabilities help accelerate development cycles and support large-scale training and testing workflows.

\begin{figure}[t]
     \centering
     \includegraphics[width=\linewidth]{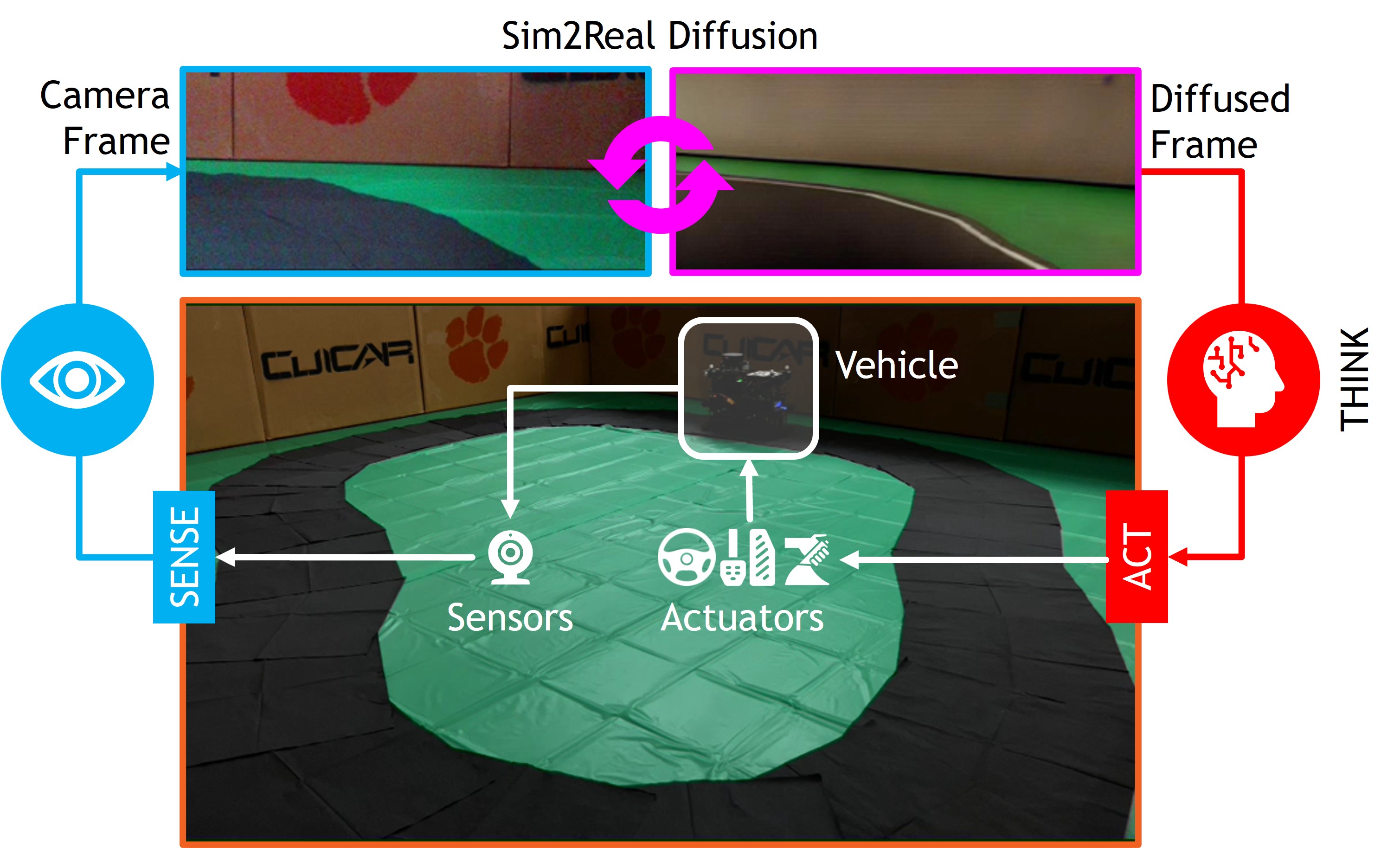}
     \caption{Proposed sim2real diffusion approach depicting the vehicle performing cross-domain perceptual adaptation for transferring simulation-trained algorithms to reality.}
     \label{fig1}
\end{figure}

However, despite all the benefits simulation has to offer, it is important to note that performance in simulation does not necessarily translate to success in the real world due to the inherent simulation-to-reality (sim2real) gap. Oftentimes, autonomy algorithms that operate flawlessly in simulation suffer from performance degradation when exposed to real-world variability and uncertainty. This discrepancy results either at the perception interface (e.g., camera images not photorealistic enough) or the dynamics interface (e.g., vehicle dynamics not accurate enough). In this work, we address the dynamics gap through careful calibration of vehicle digital twins, while primarily focusing on the visual perceptual domain gap between simulation and reality.

Previous research has tried to address the problem of sim2real
transfer through identification, adaptation, or augmentation of source/target domains, as we will review below.

\begin{table*}[t]
\centering
\caption{\small Qualitative Comparison of Proposed Method with State-of-the-Art Baselines}
\label{tab1}
\resizebox{\textwidth}{!}{%
\begin{tabular}{l|l|l|l|l|l|l|l|l|l|l}
    \hline
    \textbf{Approach}  &  \textbf{Real Data}  & \textbf{Sample Efficiency} & \textbf{Conditioning} & \textbf{ADS Decoupling} & \textbf{Stability} & \textbf{Quality} & \textbf{Modularity}  & \textbf{Generalizability} & \textbf{Reliability} & \textbf{Real-Time}  \\ \hline
    Surface Reconstruction      & Required     & Good    & N/A     & Yes & V. Good  & Good    & Yes   & V. Poor       & V. Good  & V. Good \\
    Photogrammetry              & Required     & Good    & N/A     & Yes & Good     & Good    & Yes   & V. Poor       & V. Good  & V. Good \\ 
    NeRF                        & Required     & Poor    & N/A     & Yes & Good     & V. Good & Yes   & V. Poor       & V. Good  & Poor    \\
    3DGS                        & Required     & Poor    & N/A     & Yes & Good     & V. Good & Yes   & V. Poor       & V. Good  & V. Good \\
    Domain Randomization        & Not Required & Poor    & Partial & No  & Poor     & N/A     & No    & Good          & Poor     & Good    \\
    CycleGAN                    & Required     & V. Poor & No      & Yes & Poor     & Good    & No    & Mode Collapse & Good     & Good    \\
    Transfer Learning           & Required     & Good    & No      & No  & Poor     & N/A     & No    & Poor          & Good     & Good    \\
    Curriculum Learning         & Optional     & Poor    & Partial & No  & Good     & N/A     & No    & Good          & Good     & Good    \\
    Sim2Real Diffusion (Ours)   & Not Required & V. Good & Yes     & Yes & V. Good  & V. Good & Yes   & V. Good       & V. Good  & Poor    \\ \hline
    \multicolumn{11}{l}{\footnotesize \textbf{V. Poor:} frequent failure or inefficient; \textbf{Poor:} subpar but functional; \textbf{Good:} satisfactory baseline-level performance; \textbf{V. Good:} consistently superior performance.}
\end{tabular}
}
\end{table*}

\subsection{Domain Identification}

Domain identification methods aim to discover critical parameters of the target domain (reality) and calibrate the source domain (simulation) to match those parameters. Structured identification techniques, such as surface reconstruction (SR) \cite{PoissonSurfaceRecon2006}, are geometrically precise but can falter when processing noisy or incomplete datasets and cannot generally capture visual data. Photogrammetry \cite{7780814} can closely replicate both geometric and visual details but requires calibrated image acquisition and may be less effective when applied to large-scale scenes or objects. Finally, volumetric rendering techniques such as neural radiance fields (NeRFs) \cite{NeRF2021, NeuRAD} or 3D Gaussian splatting (3DGS) \cite{kerbl20233dgaussiansplattingrealtime, SplatAD} can generate highly photorealistic representations. However, these methods are highly data-dependent and may suffer from coverage loss.

\subsection{Domain Augmentation}

Domain augmentation methods seek to inflate the source domain (simulation), which is then expected to overlap the target domain (reality) distribution. One of the most popular methods of this class is domain randomization (DR) \cite{8202133, 9294396}, which generates a wide variety of synthetic training data by randomly varying domain-specific visual attributes. Similarly, style transfer using generative adversarial networks (GANs) \cite{8793668, pmlr-v80-hoffman18a, 8620258, Stocco2023, Stocco2025} applies the style of the target domain (reality) to the source domain (simulation) during data generation.

\subsection{Domain Adaptation}

Domain adaptation usually refers to refining algorithms trained/tuned in the source domain (simulation) to perform well in the target domain (reality) with relatively less effort. Transfer learning \cite{8014892, 9341538} is one of the most commonly used methods for domain adaptation, which enables learning major differences in feature representations. However, this method offers limited flexibility (dissimilar transfer can harm performance) and can overfit to new data. Another commonly used approach is curriculum learning \cite{10.5555/3535850.3535983, 9530254}, which sequentially complicates the domain so that the model can gradually learn to adapt to the statistical differences. However, defining the learning curriculum may not always be straightforward and may introduce bias if it is ill-defined.

\subsection{Research Gaps and Contributions}

Despite several attempts, generalizable sim2real transfer still stands as an unsolved problem. Domain identification methods can potentially capture temporally and semantically consistent target-domain representations. However, one of the key limitations of these approaches is their reliance on the knowledge/access to the target domain a priori, which may not always be the case. Augmentation methods usually require additional efforts only within the source domain (simulation). However, it is difficult to capture the target domain (reality) distribution in simulation with explicit certainty, and hence, the resultant algorithms cannot be guaranteed to work well in reality. Domain adaptation offers benefits such as low-effort training in simulation (source domain) and data-driven adaptation to the real world (target domain). However, these approaches are data-dependent and require additional effort post-training to achieve reliable sim2real transfer.

In such a milieu, we try to bridge the visual perceptual domain gap by combining the strengths of augmentation and adaptation methods (refer Table \ref{tab1} for qualitative comparison with state-of-the-art). Particularly, we propose a novel sim2real transfer method for vision-based autonomy algorithms (refer Fig. \ref{fig1}) that generates an adapted camera frame representation from the target domain (real world), which closely aligns with the source domain (simulation) distribution. Our approach is designed to meet key autonomy-oriented sim2real transfer requirements such as: (i) semantically and geometrically consistent cross-domain mapping, (ii) rapid and sample-efficient generalization to novel domains, (iii) robust and modular handling of diverse sets of source and target domains, and (iv) real-time operation for closed-loop autonomy. Unlike existing sim2real transfer methods, our method offers a flexible and modular sim2real adapter, which decouples the core autonomy algorithm from the cross-domain adaptation problem, thereby significantly reducing dependency on large datasets (for retraining) or multiple models (for different domains).

The key contributions of this paper are summarized below:

\begin{itemize}
    \item \textbf{Sim2Real Diffusion:} A conditional latent diffusion pipeline capable of mapping simulation features onto real-world camera frames at runtime has been presented. The said pipeline preserves semantic relations across domains via controlled conditioning, and allows sample-efficient generalization to new domains using concept-specific \textit{``trigger words''} for modular adaptation.
    \item \textbf{Evaluation Protocol:} Selection criteria for model architecture, hyperparameters, prompt engineering, and fine-tuning have been suggested by conducting systematic ablations on the proposed framework and analyzing the effect of its individual components. Additionally, performance benchmarks are also presented across a range of computational platforms to comment on the computational feasibility of the proposed framework.
    \item \textbf{Case Studies:} Two behavioral cloning case studies across different vehicle scales and environment domains are presented. The proposed framework projects the real-time camera feed from the real-world domain back onto the simulation manifold through sim2real diffusion. The behavioral cloning models trained on simulation-only data run inference on the diffused frames that bridge the sim2real gap. Additionally, comparisons are also presented w.r.t. state-of-the-art (SOTA) baselines to position our work within existing literature.
\end{itemize}


\section{Research Methodology}
\label{Section: Research Methodology}

\begin{figure*}[t]
     \centering
     \includegraphics[width=\linewidth]{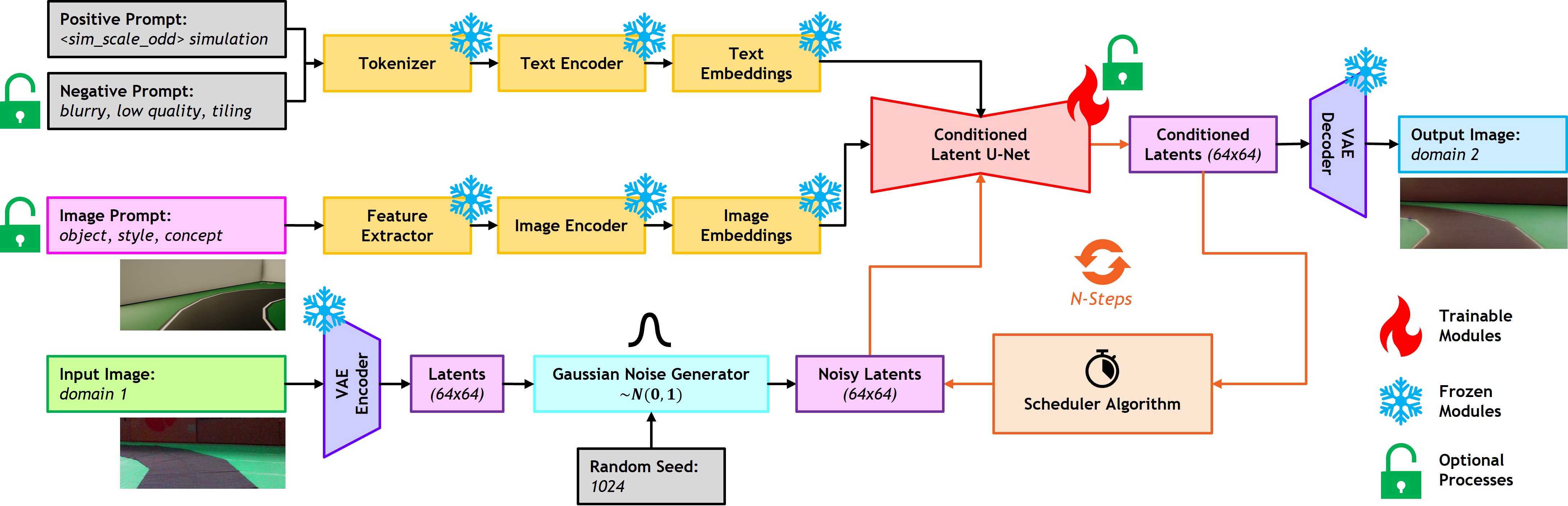}
     \caption{Proposed framework for enabling sim2real transfer of vision-based autonomy algorithms through the learning of adaptive cross-domain representations using image and text conditioning within a latent diffusion architecture.}
     \label{fig2}
\end{figure*}

\subsection{Sim2Real Diffusion}

Diffusion probabilistic models (DPMs) learn ``denoising'' by reversing the diffusion process, a Markov chain of length $T$ that incrementally adds noise to the data. This effectively trains the model to learn the original data distribution $p(x)$ by removing Gaussian noise over multiple steps. These models can be viewed as a sequence of denoising autoencoders $\epsilon_\theta(x_t, t)$, each trained to recover cleaner inputs from progressively noisier ones. The training objective is simplified by varying the noise step $t$ uniformly, i.e., $t = \{1, ..., T\}$:
\begin{align}
\label{DPM-Loss}
\mathcal{L}_\textrm{DPM} = \mathbb{E}_{x, \epsilon \sim \mathcal{N}(0,1), t} \left[ \left\| \epsilon - \epsilon_\theta(x_t, t) \right\|_2^2 \right]
\end{align}

The proposed sim2real diffusion framework (refer Fig. \ref{fig2}) leverages conditional latent diffusion models (LDM) \cite{LDM2022}, which operate in a latent space derived using an autoencoder. Particularly, the input image $x \in \mathbb{R}^{H \times W \times 3}$, represented in RGB format, is transformed by the encoder $\mathcal{E}$ into a latent vector $z = \mathcal{E}(x)$. This latent representation $z \in \mathbb{R}^{h \times w \times c}$ is later passed through the decoder $\mathcal{D}$, which reconstructs the image $\tilde{x} = \mathcal{D}(z) = \mathcal{D}(\mathcal{E}(x))$. In the context of LDMs, Eq. \ref{DPM-Loss} can be rewritten to include the latent representation:
\begin{align}
\label{LDM-Loss}
\mathcal{L}_\textrm{LDM} = \mathbb{E}_{\mathcal{E}(x), \epsilon \sim \mathcal{N}(0,1), t} \left[ \left\| \epsilon - \epsilon_\theta(z_t, t) \right\|_2^2 \right]
\end{align}

The core neural architecture $\epsilon_\theta(\circ, t)$ of the diffusion model is a time-conditioned U-Net \cite{UNet2015}, comprising ResNet-based encoder and decoder blocks. The encoder downsamples the latent image, while the decoder reconstructs a cleaner version by predicting the noise residual, which is then used to compute a denoised latent image representation via a scheduler algorithm (e.g., DPM-Solver). The corresponding layers of the encoder and decoder are linked via skip connections to preserve critical features. Since the forward diffusion process is predetermined, the latent variable $z_t$ can be efficiently derived from the encoder $\mathcal{E}$ during training. Similarly, a single pass through the decoder $\mathcal{D}$ is sufficient to transform samples from the latent space $p(z)$ back into the image space.

Controlling the generative process of latent diffusion models through auxiliary inputs $y$ such as text prompts, images, semantic maps, depth maps, etc., is possible, but requires modeling conditional probability distributions of the form $p(z|y)$. This can be realized with a conditional denoising autoencoder $\epsilon_\theta(z_t, t, y)$, which in the context of U-Net, can be implemented via the cross-attention mechanism. Particularly, a domain-specific encoder $\tau_\theta$ is used to project multi-modal condition $y$
onto an intermediate manifold $\tau_\theta(y) \in \mathbb{R}^{M \times d_\tau}$, which is then injected between the ResNet blocks of the U-Net via the cross-attention mechanism:
\begin{align}
\label{Cross-Attention}
\texttt{Attention}(Q, K, V) = \texttt{Softmax}\left( \frac{Q \cdot K^\top}{\sqrt{d}} \right) \cdot V
\end{align}
where, $Q = W_Q^{(i)} \cdot \varphi_i(z_t)$, $K = W_K^{(i)} \cdot \tau_\theta(y)$, $V = W_V^{(i)} \cdot \tau_\theta(y)$ such that $W_Q^{(i)} \in \mathbb{R}^{d \times d_\tau}$, $W_K^{(i)} \in \mathbb{R}^{d \times d_\tau}$, $W_V^{(i)} \in \mathbb{R}^{d \times d_\epsilon^i}$ are learnable projection matrices, and $\varphi_i(z_t) \in \mathbb{R}^{N \times d_\epsilon^i}$ denotes a flattened intermediate representation of U-Net $\epsilon_\theta$.

In the context of conditional LDMs, Eq. \ref{LDM-Loss} can be rewritten to include the condition $y$ by jointly optimizing $\tau_\theta$ and $\epsilon_\theta$:
\begin{align}
\label{C-LDM-Loss}
\mathcal{L}_\textrm{C-LDM} = \mathbb{E}_{\mathcal{E}(x), y, \epsilon \sim \mathcal{N}(0,1), t} \left[ \left\| \epsilon - \epsilon_\theta(z_t, t, \tau_\theta(y)) \right\|_2^2 \right]
\end{align}

Additionally, the proposed pipeline also supports IP-Adapter \cite{IPAdapter2023}, which employs a decoupled cross-attention mechanism for separating the text $y_\textrm{txt}$ and image $y_\textrm{img}$ features. Consequently, we have an additional term in Eq. \ref{Cross-Attention}, corresponding to the image prompt:
\begin{align}
\label{Decoupled-Cross-Attention}
\texttt{Softmax}\left( \frac{Q \cdot K^\top}{\sqrt{d}} \right) \cdot V + \texttt{Softmax}\left( \frac{Q \cdot (K')^\top}{\sqrt{d}} \right) \cdot V'
\end{align}
where, $Q = W_Q^{(i)} \cdot \varphi_i(z_t)$, $K = W_K^{(i)} \cdot \tau_\theta(y_\textrm{txt})$, $V = W_V^{(i)} \cdot \tau_\theta(y_\textrm{txt})$, $K' = W_K'^{(i)} \cdot \tau_\theta(y_\textrm{img})$, and $V' = W_V'^{(i)} \cdot \tau_\theta(y_\textrm{img})$.

In this context, Eq. \ref{C-LDM-Loss} can be rewritten to include the IP-Adapter:
\begin{multline}
\label{IPC-LDM-Loss}
\mathcal{L}_\textrm{IPC-LDM} = \mathbb{E}_{\mathcal{E}(x), y_\textrm{txt}, y_\textrm{img}, \epsilon \sim \mathcal{N}(0,1), t} \\ \left[ \left\| \epsilon - \epsilon_\theta(z_t, t, \tau_\theta(y_\textrm{txt}), \tau_\theta(y_\textrm{img})) \right\|_2^2 \right]
\end{multline}

The conditional prompts are encoded into an embedding space that can be understood by the U-Net. Text prompts are encoded into latent text embeddings by a transformer-based encoder (e.g., CLIPTextModel). Similarly, image prompts are encoded into patch image embeddings by a vision transformer-based encoder (e.g., OpenCLIP-ViT-H-14). Here, positive prompts direct the generative process by providing context and instructions. The proposed framework leverages positive prompts with innovative concept-specific \textit{``trigger words''} of the form \texttt{<sim\_scale\_odd>} to modularly identify/associate a particular concept with a particular simulator, scale, and operational design domain (ODD). Negative prompts describe undesired attributes, which enhances the generative process by explicitly telling the model what to avoid in the output. The image prompt (IP) offers valuable information in the form of a reference concept/style, which guides the generative process -- the output is expected to have the same semantics as the input image, but with the style of the prompt image. The difference between input and style is made clear to the U-Net by using a dedicated encoder for the image prompt, which is independent of the input encoder. This follows that there is no pair-wise relationship between the input image(s) that come from the real world and the prompt image that is a single sample from the simulation.

\subsection{Design of Experiments}

\begin{figure*}[t]
     \centering
     \begin{subfigure}[b]{0.16\linewidth}
         \centering
         \includegraphics[width=\linewidth]{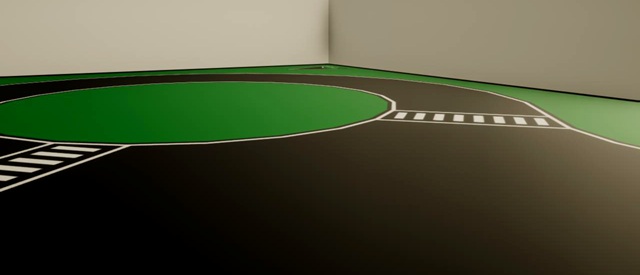}
         \caption{}
         \label{fig3a}
     \end{subfigure}
     \hfill
     \begin{subfigure}[b]{0.16\linewidth}
         \centering
         \includegraphics[width=\linewidth]{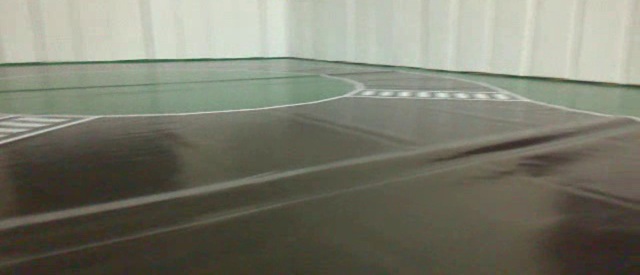}
         \caption{}
         \label{fig3b}
     \end{subfigure}
     \hfill
     \begin{subfigure}[b]{0.16\linewidth}
         \centering
         \includegraphics[width=\linewidth]{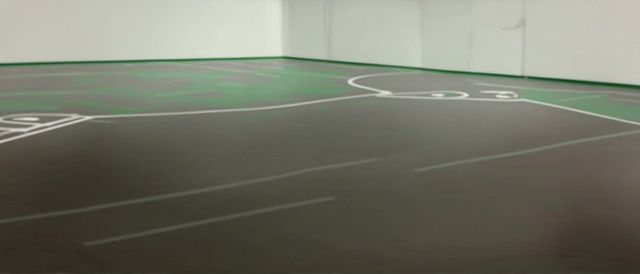}
         \caption{}
         \label{fig3c}
     \end{subfigure}
     \hfill
     \begin{subfigure}[b]{0.16\linewidth}
         \centering
         \includegraphics[width=\linewidth]{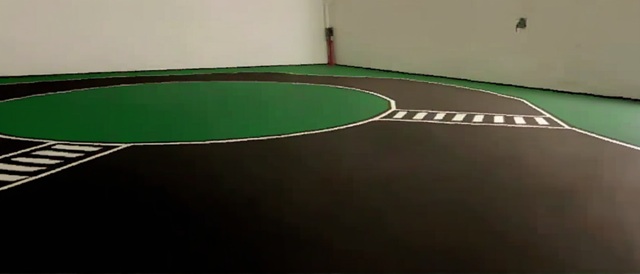}
         \caption{}
         \label{fig3d}
     \end{subfigure}
     \hfill
     \begin{subfigure}[b]{0.16\linewidth}
         \centering
         \includegraphics[width=\linewidth]{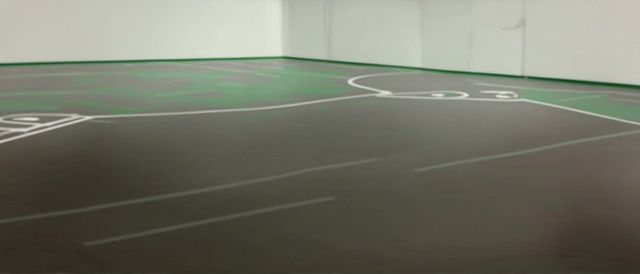}
         \caption{}
         \label{fig3e}
     \end{subfigure}
     \hfill
     \begin{subfigure}[b]{0.16\linewidth}
         \centering
         \includegraphics[width=\linewidth]{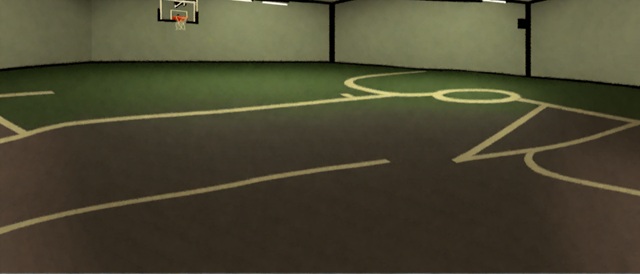}
         \caption{}
         \label{fig3f}
     \end{subfigure}
     \begin{subfigure}[b]{0.16\linewidth}
         \centering
         \includegraphics[width=\linewidth]{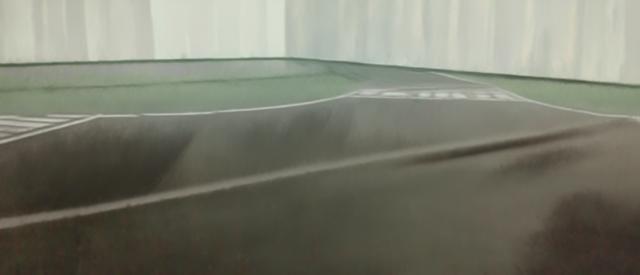}
         \caption{}
         \label{fig3g}
     \end{subfigure}
     \hfill
     \begin{subfigure}[b]{0.16\linewidth}
         \centering
         \includegraphics[width=\linewidth]{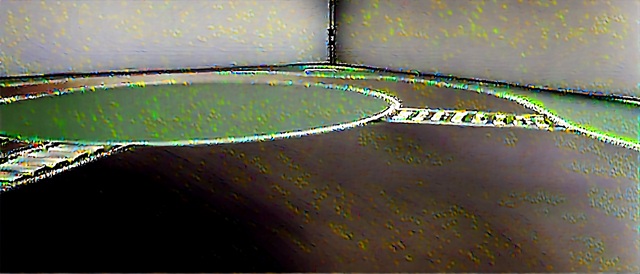}
         \caption{}
         \label{fig3h}
     \end{subfigure}
     \hfill
     \begin{subfigure}[b]{0.16\linewidth}
         \centering
         \includegraphics[width=\linewidth]{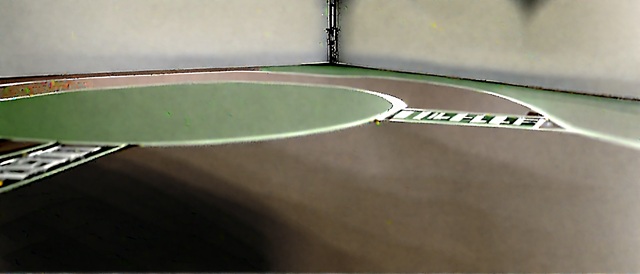}
         \caption{}
         \label{fig3i}
     \end{subfigure}
     \hfill
     \begin{subfigure}[b]{0.16\linewidth}
         \centering
         \includegraphics[width=\linewidth]{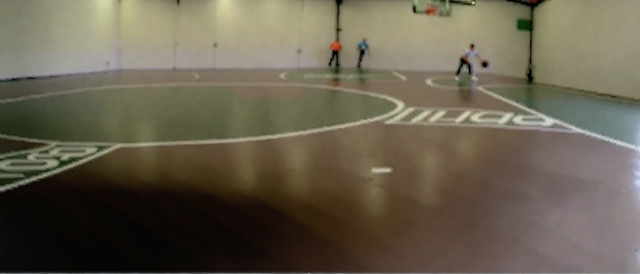}
         \caption{}
         \label{fig3j}
     \end{subfigure}
     \hfill
     \begin{subfigure}[b]{0.16\linewidth}
         \centering
         \includegraphics[width=\linewidth]{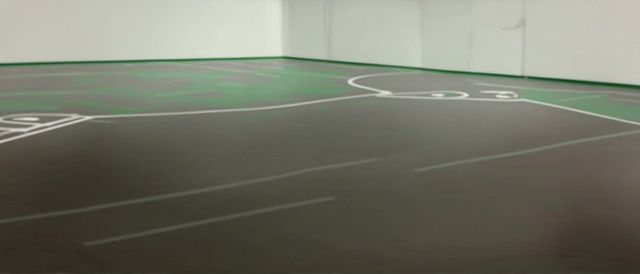}
         \caption{}
         \label{fig3k}
     \end{subfigure}
     \hfill
     \begin{subfigure}[b]{0.16\linewidth}
         \centering
         \includegraphics[width=\linewidth]{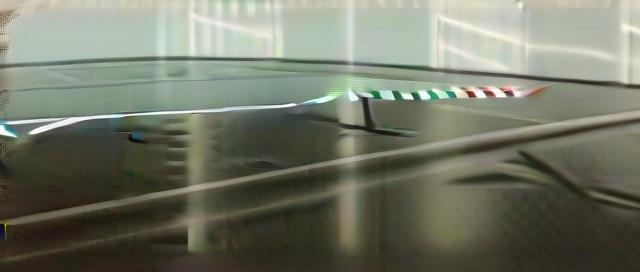}
         \caption{}
         \label{fig3l}
     \end{subfigure}
     \caption{Results of the 1$^\textrm{st}$ ablation study performed on (a) sim and (b) real camera frames to assess the effect of domain adaptation direction \{(c) real2sim, (d) sim2real\}, foundation model \{(e) SDXL, (f) SDXL-Turbo, (g) SD3M\}, denoising steps \{(h) 1, (i) 5, (j) 30\}, and input image resolution \{(k) 2560×1096 px, (l) 640×274 px\}.}
     \label{fig3}
\end{figure*}

\begin{figure*}[t]
     \centering
     \begin{subfigure}[b]{0.16\linewidth}
         \centering
         \includegraphics[width=\linewidth]{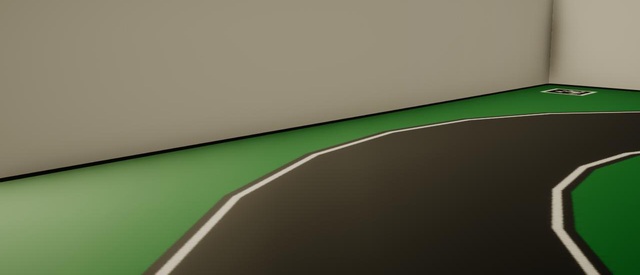}
         \caption{}
         \label{fig4a}
     \end{subfigure}
     \hfill
     \begin{subfigure}[b]{0.16\linewidth}
         \centering
         \includegraphics[width=\linewidth]{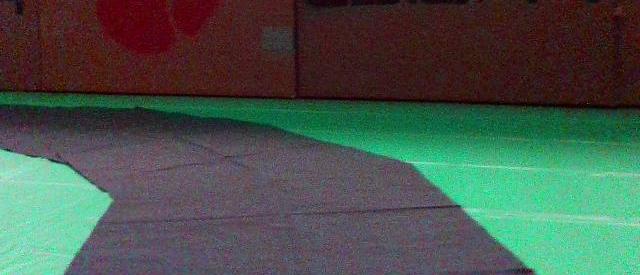}
         \caption{}
         \label{fig4b}
     \end{subfigure}
     \hfill
     \begin{subfigure}[b]{0.16\linewidth}
         \centering
         \includegraphics[width=\linewidth]{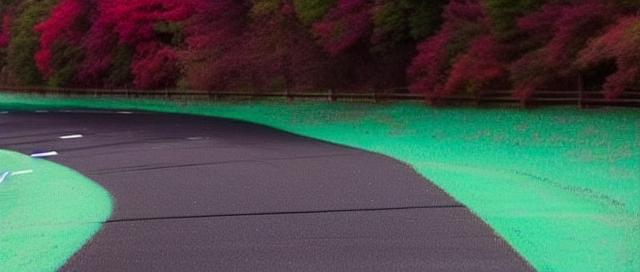}
         \caption{}
         \label{fig4c}
     \end{subfigure}
     \hfill
     \begin{subfigure}[b]{0.16\linewidth}
         \centering
         \includegraphics[width=\linewidth]{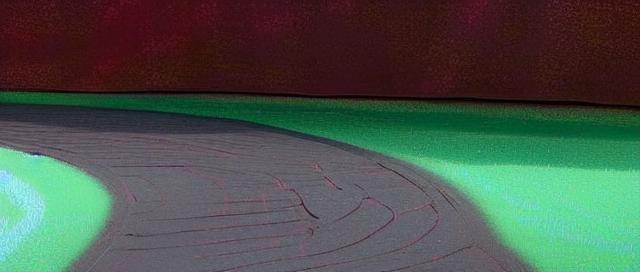}
         \caption{}
         \label{fig4d}
     \end{subfigure}
     \hfill
     \begin{subfigure}[b]{0.16\linewidth}
         \centering
         \includegraphics[width=\linewidth]{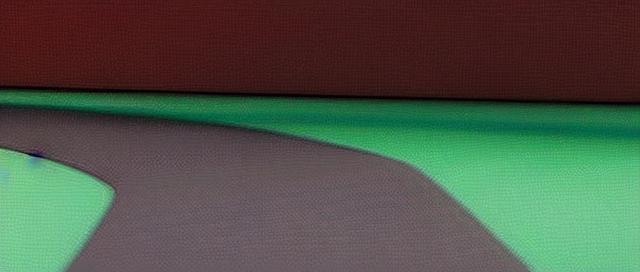}
         \caption{}
         \label{fig4e}
     \end{subfigure}
     \hfill
     \begin{subfigure}[b]{0.16\linewidth}
         \centering
         \includegraphics[width=\linewidth]{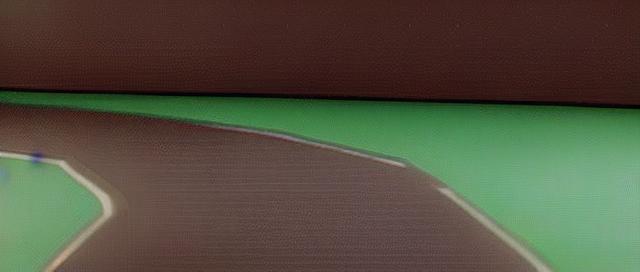}
         \caption{}
         \label{fig4f}
     \end{subfigure}
     \caption{Results of the 2$^\textrm{nd}$ ablation study performed on model architecture with a common (a) image prompt (if applicable) and (b) input image, to assess the effect of image-prompting \{(c, d) no, (e, f) yes\} and fine-tuning \{(c, e) no, (d, f) yes\}.}
     \label{fig4}
\end{figure*}

All the experiments in this work were conducted using AutoDRIVE Ecosystem \cite{AutoDRIVE2023}. Particularly, AutoDRIVE Simulator was employed to simulate the autonomy-oriented digital twins of Nigel (small-scale, on-road), RoboRacer (small-scale, racing), and Hunter/RZR (large-scale, off-road) to collect training and testing data for the various ablation studies. Additionally, AutoDRIVE Testbed (with Nigel/Hunter) was used for the end-to-end autonomous driving case studies.

\subsubsection{Research Questions}

The goal of the design of experiments is to guide the investigation of the following research questions: (RQ1) What are the desired hyperparameters and model backbones for effective sim2real diffusion? (RQ2) What are the desired conditioning and fine-tuning configurations for sim2real diffusion? (RQ3) Can the same framework generalize to exaggerated domain gaps via conditioning? (RQ4) Can the framework conveniently handle multiple domain representations via modular fine-tuning? (RQ5) Can the framework remain scalable and servicable to the broader community? (RQ6) Can the framework handle runtime domain adaptation during real-world vehicle deployments? and (RQ7) How does the sim2real diffusion perform against comparable baseline approaches?

\subsubsection{Ablation Studies}

We performed ablation studies in 4 stages. The first set of experiments was meant to assess the effect of (i) domain adaptation direction, (ii) foundation model, (iii) denoising steps, and (iv) input resolution to answer RQ1. The second, third, and fourth sets of experiments were meant to assess the effect of (i) fine-tuning and (ii) image prompting, qualitatively and quantitatively, to answer RQ2, RQ3, and RQ4, respectively. To this end, we fine-tuned the SD1.5 foundation model on 3 custom concepts using the DreamBooth approach \cite{DreamBooth2023}. The fine-tuning comprised a \textit{``trigger word''} of the form \texttt{<sim\_scale\_odd>}, enabling sample-efficient fine-tuning (FT) and modular adaptation across multiple domains with minimal effort.

\subsubsection{Performance Study}

We analyzed the framework across a range of computational platforms to answer RQ5. This involved training and inference of the sim2real diffusion model on (i) Palmetto Cluster with 4 different GPU models \{V100, A100, H100, L40S\}, (ii) Google Colab with T4 GPU, (iii) laptop PC with 3080 Ti GPU, and edge compute with 2060 GPU.

\subsubsection{Case Studies}

Although our sim2real transfer approach is decoupled and not limited to the model used for (or any aspect of the) ADS, the chosen case studies attempt to investigate RQ6 through behavioral cloning examples in \cite{RBC2021, dave2}, but with 2 key differences, which exaggerated the sim2real gap: (i) no data augmentation during training, and (ii) deployment in a dissimilar real-world environment. Particularly, AutoDRIVE Simulator was employed for collecting $\sim$5 laps worth of manual driving data ($\sim$1700 samples for Nigel and $\sim$3000 samples for Hunter), which was balanced, normalized, and resized. Two deep convolutional neural network (CNN) models (ADSs) were trained while leveraging a custom architecture by Samak et al. in \cite{RBC2021} for Nigel (to mimic experimental protocol) and DAVE-2 by Bojarski et al. in \cite{dave2} for Hunter (to avoid selection bias) while keeping other aspects of the pipeline the same. For Nigel, the simulation environment was an indoor figure-8 track comprising a black road with white lane markings on a green vinyl mat surrounded by white boxes. Contrarily, the real-world environment for Nigel consisted of a rather irregular track created using black tissues (different textures, flat edges, folds) laid out on a green tablecloth (with folds, creases, reflections) surrounded by brown cardboard boxes (with imprints, tapes, damage). Additionally, the camera model in simulation (Raspberry Pi Camera V2) and reality (Intel RealSense D435i) were different, resulting in a perception gap due to film grain, exposure, resolution, and field of view (FOV) of the perceived image. For Hunter, we kept the camera setup the same and only varied the environment from an outdoor dirt road in simulation to straight (S) and curved (C) sidewalks, potentially with obstacles (SO/CO) in the real world.

\subsubsection{Evaluation Metrics}

Drawing inspiration from Lambertenghi, et al. \cite{Stocco2024}, the proposed sim2real diffusion framework was assessed across the following metrics: cosine similarity (CS) of the input and output image embeddings; CLIP score (CLIP-S) and CLIP directional similarity (CLIP-DS) between embeddings of input and output image/caption pairs; learned perceptual image patch similarity (LPIPS) between input and output images; Gram-matrix style difference (SD) between the input-style pair (SD-IS) and output-style pair (SD-OS); Fr\'echet inception distance (FID), kernel inception distance (KID), and inception score (IS) between features of input/style and output images; and the performance of the proposed framework was measured based on the denoising iterations per second (IPS) analogous to frequency in Hz. Additionally, training time (TT), average progress along track (PAT), root mean squared cross-track error (RMS-XTE), and success rate (SR) were employed to investigate RQ7.


\section{Results and Discussion}
\label{Section: Results and Discussion}

\subsection{Ablation Studies}

The first ablation study (refer Fig. \ref{fig3} and Table \ref{tab2}) was meant to assess the fitness of latent diffusion models for domain adaptation. Particularly, a common set of virtual and real camera frames (from Nigel dataset) was provided to the diffusion model(s) to assess their capabilities across the following mutually independent experiments.

\textbf{Domain Adaptation Direction:} It was observed that adapting real-world images to simulated images (a.k.a. real2sim mapping) was semantically better than sim2real mapping. This can be attributed to the fact that it involves simplifying complex, noisy inputs into cleaner, lower-variance outputs, making it a more stable and easier many-to-one mapping than the detail-rich, multimodal sim2real direction, which, being a one-to-many mapping, can face scalability issues across varied ODDs.
    
\textbf{Foundation Model:} It was observed that for the common task of sim2real mapping, the base model (SDXL) performed much better than its corresponding time-distilled counterpart (SDXL-Turbo), since time-distillation usually leads to over-creativity, which is not suitable for domain adaptation.

\textbf{Denoising Steps:} It was observed that for the task of real2sim mapping using the same model (SDXL-Turbo), denoising steps heavily influence the generative output. While too few steps (e.g., just 1) lead to noisy output with insufficient domain adaptation, too many of them (e.g., more than 30) quickly lead to hallucination (notice the basketball, hoop, and players generated by the model in Fig. \hyperref[fig3]{\ref*{fig3}(j)}).

\textbf{Input Resolution:} While the proposed pipeline is agnostic to the input image size/resolution, these values do affect the generative process (if the foundation model is originally trained on a specific resolution dataset). It was observed that with the same base model (SDXL), reducing the input resolution progressively degraded the generative output, with a significant feature loss beyond a certain threshold (640$\times$274 px, refer Fig. \hyperref[fig3]{\ref*{fig3}(l)}). It was later observed that fine-tuning could adapt the foundation model(s) to novel input resolutions.

\begin{figure}[t]
     \centering
     \begin{subfigure}[b]{0.15\linewidth}
         \centering
         \includegraphics[width=\linewidth]{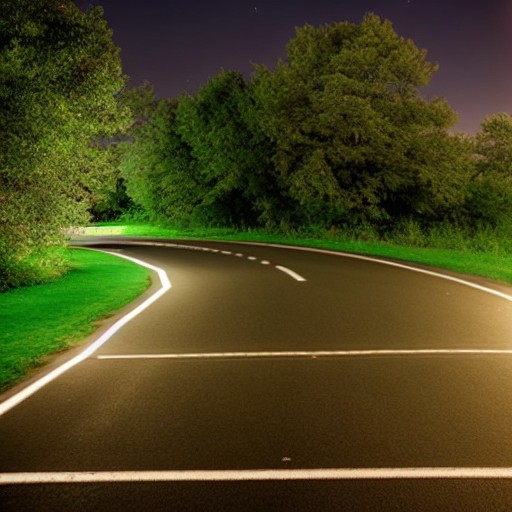}
         \caption{}
         \label{fig5a}
     \end{subfigure}
     \hfill
     \begin{subfigure}[b]{0.15\linewidth}
         \centering
         \includegraphics[width=\linewidth]{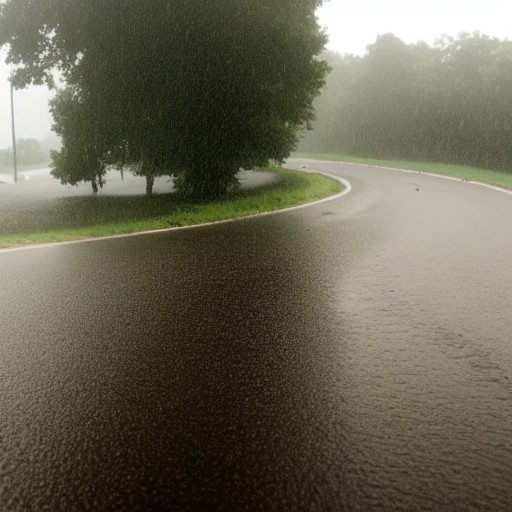}
         \caption{}
         \label{fig5b}
     \end{subfigure}
     \hfill
     \begin{subfigure}[b]{0.15\linewidth}
         \centering
         \includegraphics[width=\linewidth]{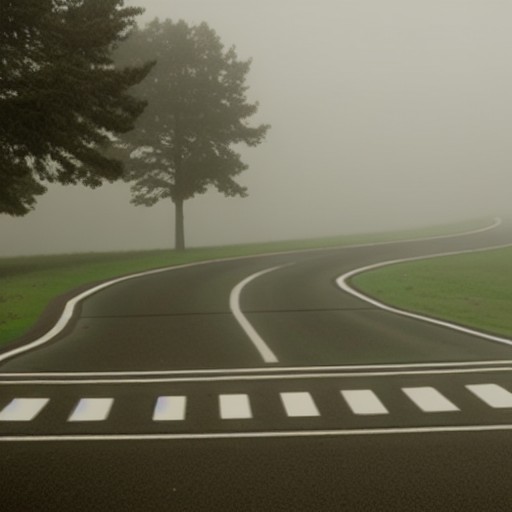}
         \caption{}
         \label{fig5c}
     \end{subfigure}
     \hfill
     \begin{subfigure}[b]{0.15\linewidth}
         \centering
         \includegraphics[width=\linewidth]{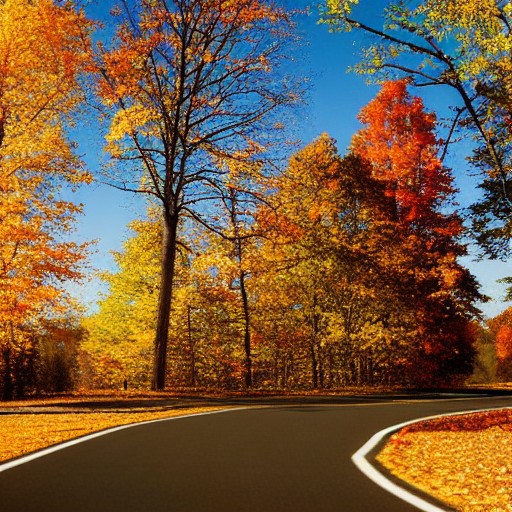}
         \caption{}
         \label{fig5d}
     \end{subfigure}
     \hfill
     \begin{subfigure}[b]{0.15\linewidth}
         \centering
         \includegraphics[width=\linewidth]{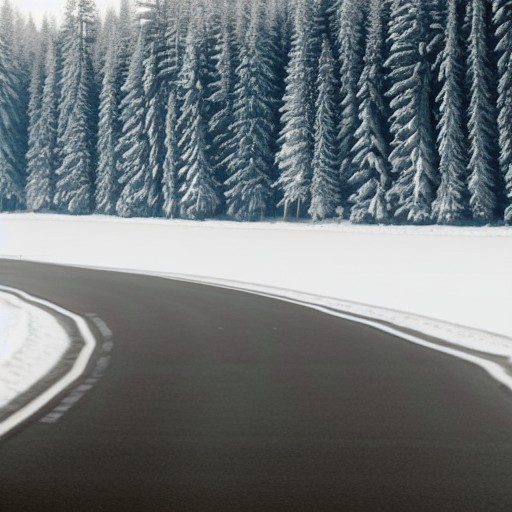}
         \caption{}
         \label{fig5e}
     \end{subfigure}
     \hfill
     \begin{subfigure}[b]{0.15\linewidth}
         \centering
         \includegraphics[width=\linewidth]{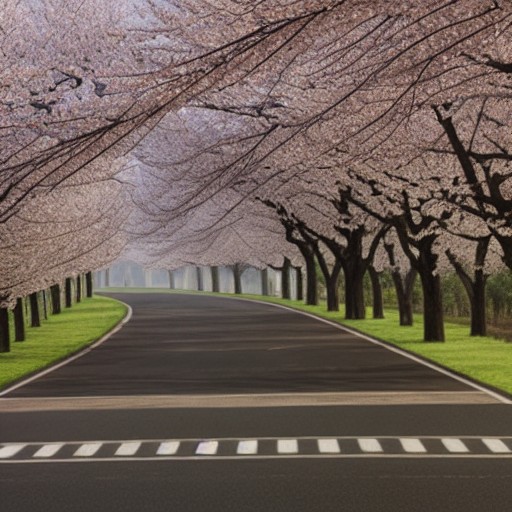}
         \caption{}
         \label{fig5f}
     \end{subfigure}
     \begin{subfigure}[b]{0.15\linewidth}
         \centering
         \includegraphics[width=\linewidth]{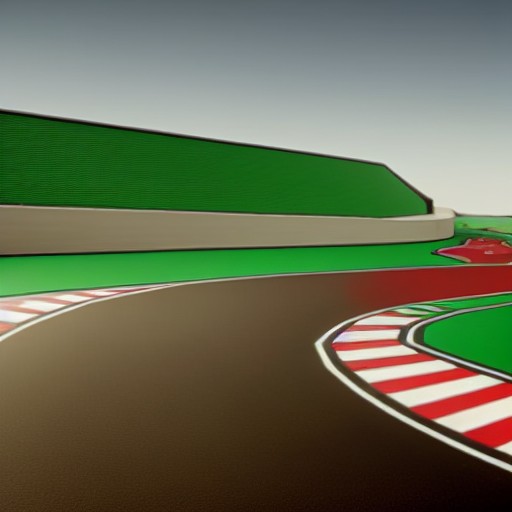}
         \caption{}
         \label{fig5g}
     \end{subfigure}
     \hfill
     \begin{subfigure}[b]{0.15\linewidth}
         \centering
         \includegraphics[width=\linewidth]{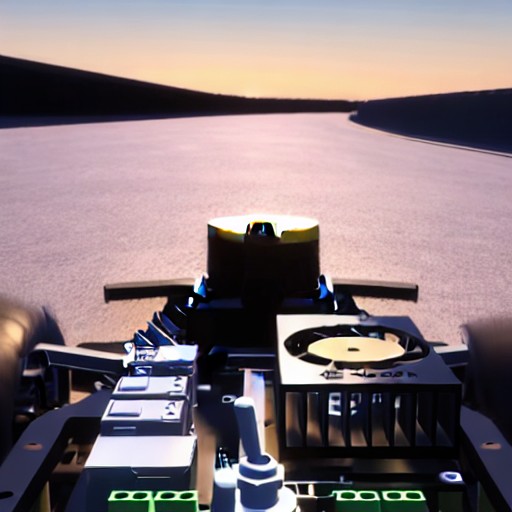}
         \caption{}
         \label{fig5h}
     \end{subfigure}
     \hfill
     \begin{subfigure}[b]{0.15\linewidth}
         \centering
         \includegraphics[width=\linewidth]{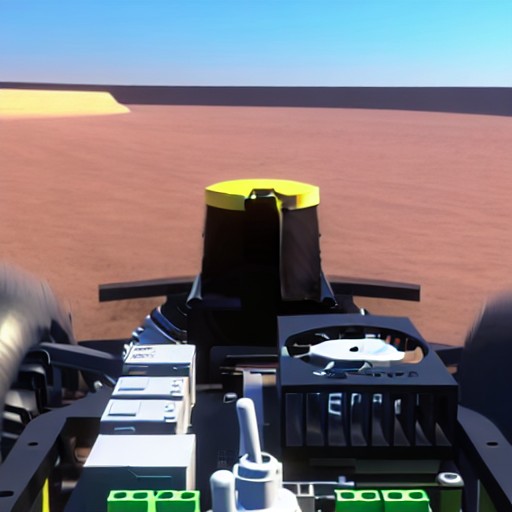}
         \caption{}
         \label{fig5i}
     \end{subfigure}
     \hfill
     \begin{subfigure}[b]{0.15\linewidth}
         \centering
         \includegraphics[width=\linewidth]{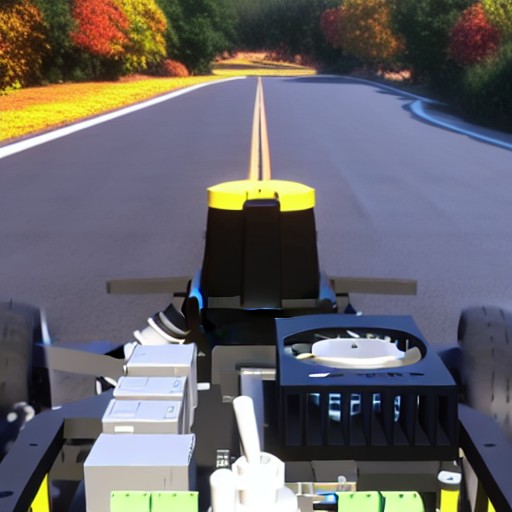}
         \caption{}
         \label{fig5j}
     \end{subfigure}
     \hfill
     \begin{subfigure}[b]{0.15\linewidth}
         \centering
         \includegraphics[width=\linewidth]{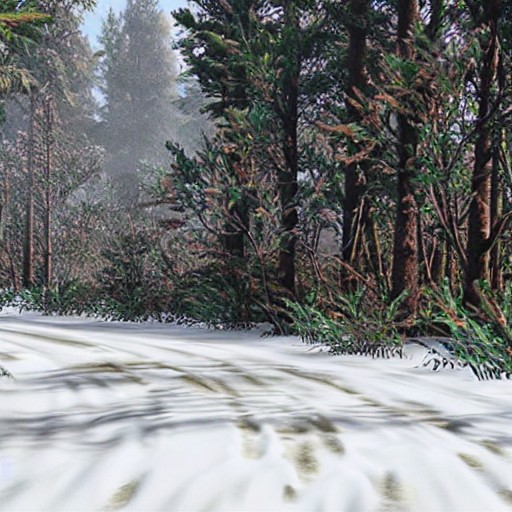}
         \caption{}
         \label{fig5k}
     \end{subfigure}
     \hfill
     \begin{subfigure}[b]{0.15\linewidth}
         \centering
         \includegraphics[width=\linewidth]{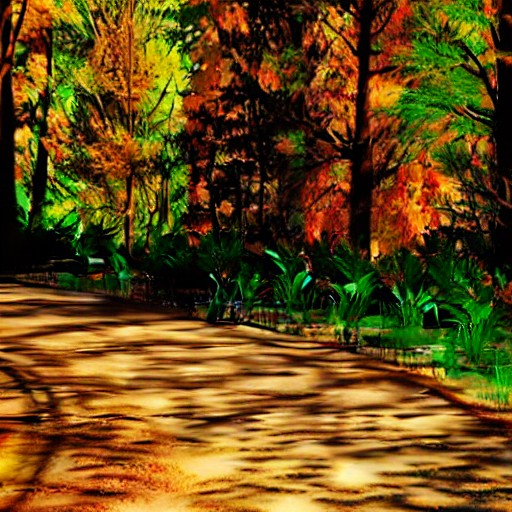}
         \caption{}
         \label{fig5l}
     \end{subfigure}
     \caption{Results of the 3$^\textrm{rd}$ ablation study performed on fine-tuned model with pure text conditioning to assess output sample diversity: \texttt{<autodrive\_small\_onroad>} (a) night, (b) rain, (c) fog, (d) fall, (e) winter, (f) spring, (g) racetrack; \texttt{<autodrive\_small\_racing>} (h) sunrise, (i) desert, (j) public road; \texttt{<autodrive\_large\_offroad>} (k) snow, (l) fall.}
     \label{fig5}
\end{figure}

\begin{table}[t]
\centering
\caption{\small Quantitative Results of Ablation Study 1}
\label{tab2}
\resizebox{\columnwidth}{!}{%
\begin{tabular}{l|l|l|l|l}
    \hline
    \textbf{Ablation}  &  \textbf{CS $\uparrow$}  & \textbf{CLIP-DS $\uparrow$} & \textbf{LPIPS $\downarrow$}  & \textbf{SD-OS$^{\mathrm{\dagger}}$ $\downarrow$}  \\ \hline
    Real $\mapsto$ Sim     & 8.40e-01    & 2.37e-02    & 2.52e-01    & 5.41e-04 \\
    Sim $\mapsto$ Real     & 7.77e-01    & 1.41e-03    & 6.01e-01    & 4.14e-04 \\ \hline
    SDXL Model             & 8.40e-01    & 2.37e-02    & 2.52e-01    & 5.41e-04 \\
    SDXL-Turbo Model       & 7.30e-01    & 1.08e-01    & 4.98e-01    & 4.03e-04 \\
    SD3M Model             & 7.24e-01    & 4.34e-02    & 1.86e-01    & 8.24e-04 \\ \hline
    1 Denoising Step       & 7.54e-01    & 5.71e-02    & 6.60e-01    & 3.21e-03 \\
    5 Denoising Steps      & 7.38e-01    & 9.89e-02    & 5.86e-01    & 3.31e-04 \\
    30 Denoising Steps     & 7.91e-01    & -1.08e-02   & 6.31e-01    & 7.95e-04 \\ \hline
    2560×1096 Resolution   & 8.40e-01    & 2.37e-02    & 2.52e-01    & 5.41e-04 \\
    640×274 Resolution     & 7.13e-01    & 4.90e-02    & 3.91e-01    & 6.41e-04 \\  \hline
    \multicolumn{5}{l}{\footnotesize $^\mathrm{\dagger}$SD-IS = 6.78e-04.}
\end{tabular}
}
\end{table}

\begin{table}[t]
\centering
\caption{\small Quantitative Results of Ablation Study 2}
\label{tab3}
\resizebox{\columnwidth}{!}{%
\begin{tabular}{l|l|l|l|l|l}
    \hline
    \textbf{Model}  &  \textbf{CS $\uparrow$}  & \textbf{CLIP-DS $\uparrow$} & \textbf{LPIPS $\downarrow$}  & \textbf{SD-OS$^{\mathrm{\dagger}}$ $\downarrow$}    & \textbf{IPS$^{\mathrm{\ddagger}}$ $\uparrow$}   \\ \hline
    U-Net       & 7.04e-01    & 6.10e-02    & 3.74e-01    & 9.17e-04    & 7.36    \\
    U-Net+IP    & 7.58e-01    & 1.17e-01    & 3.43e-01    & 7.80e-04    & 6.73    \\
    U-Net+FT         & 8.05e-01    & 1.05e-01    & \textbf{2.64e-01}    & 7.53e-04    & \textbf{7.42}    \\
    U-Net+FT+IP      & \textbf{8.19e-01}    & \textbf{1.78e-01}    & 2.76e-01    & \textbf{6.89e-04}    & 6.86    \\ \hline
    \multicolumn{6}{l}{\footnotesize $^\mathrm{\dagger}$SD-IS = 1.24e-03, $^\mathrm{\ddagger}$Laptop GPU: 3080 Ti.}
\end{tabular}
}
\end{table}

The second ablation study (refer Fig. \ref{fig4} and Table \ref{tab3}) was performed to assess the effects of fine-tuning and image-prompting on latent diffusion models. Similar to the earlier study, a common set of input (real) and prompt (sim) frames were provided to the diffusion model(s) to assess their domain adaptation capabilities across a range of experiments.

\textbf{Image-Prompting:} Image prompt (IP) provides additional conditioning, which significantly aids in domain adaptation (compare Fig. \hyperref[fig4]{\ref*{fig4}(c)} with Fig. \hyperref[fig4]{\ref*{fig4}(e)}, or Fig. \hyperref[fig4]{\ref*{fig4}(d)} with Fig. \hyperref[fig4]{\ref*{fig4}(f)}). It was observed that, as opposed to the vanilla models, those conditioned via an image prompt performed better across all the quantitative metrics (compare Table \ref{tab3} rows 1 and 2, or 3 and 4), albeit adding a small processing overhead.
    
\textbf{Fine-Tuning:} Fine-tuning the foundation model for a custom concept serves a dual purpose. Firstly, it improves the qualitative (compare Fig. \hyperref[fig4]{\ref*{fig4}(c)} with Fig. \hyperref[fig4]{\ref*{fig4}(d)}, or Fig. \hyperref[fig4]{\ref*{fig4}(e)} with Fig. \hyperref[fig4]{\ref*{fig4}(f)}) and quantitative (compare Table \ref{tab3} rows 1 and 3, or 2 and 4) performance of domain adaptation. Secondly, fine-tuning using a \textit{``trigger word''} for each concept makes this approach highly modular and scalable across novel domain gaps. Finally, combining fine-tuning with image-prompting usually results in the most effective domain adaptation, as marked by the highest (44.41\%) reduction in style difference.

The third ablation study (refer Fig. \ref{fig5} and Table \ref{tab4}) served the purpose of assessing the scalability/generalizability across exaggerated domain gaps. To this end, the latent diffusion model fine-tuned on 3 different concepts (on-road, racing, and off-road) was stress-tested for its text-to-image generative creativity. Fig. \ref{fig5} shows that the fine-tuned model is able to map domains across various weathers, seasons, and times of the day, and is also capable of mapping one ODD to another (refer Fig. \hyperref[fig5]{\ref*{fig5}(g)} or Fig. \hyperref[fig5]{\ref*{fig5}(j)}) or altering the ODD altogether (refer Fig. \hyperref[fig5]{\ref*{fig5}(i)}), simply using text prompts. Consequently, unlike other ablations, the metrics reported in Table \ref{tab4} focus on quantifying the model's capability to generate diverse samples in text-to-image tasks.

\begin{table}[t]
\centering
\caption{\small Quantitative Results of Ablation Study 3}
\label{tab4}
\resizebox{\columnwidth}{!}{%
\begin{tabular}{l|l|l|l|l|l}
    \hline
    \textbf{FT Domain}  &  \textbf{FID $\downarrow$} &  \textbf{KID $\downarrow$}  & \textbf{CLIP-S $\uparrow$} & \textbf{LPIPS $\uparrow$}  &  \textbf{IS $\uparrow$}   \\ \hline
    Small, On-Road       & 53.68    & 2.41e-01    & 3.58e-01   & 5.21e-01    & 8.64   \\
    Small, Racing        & 27.76    & 7.26e-02    & 2.36e-01   & 3.14e-01    & 5.21   \\
    Large, Off-Road      & 38.52    & 1.83e-01    & 3.22e-01   & 4.72e-01    & 7.92   \\ \hline
\end{tabular}
}
\end{table}

\begin{figure}[t]
     \centering
     \begin{subfigure}[b]{0.32\linewidth}
         \centering
         \includegraphics[width=\linewidth]{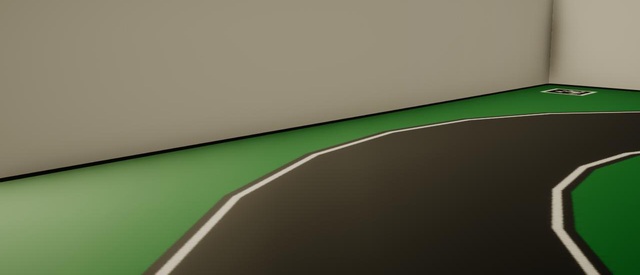}
         \caption{}
         \label{fig6a}
     \end{subfigure}
     \hfill
     \begin{subfigure}[b]{0.32\linewidth}
         \centering
         \includegraphics[width=\linewidth]{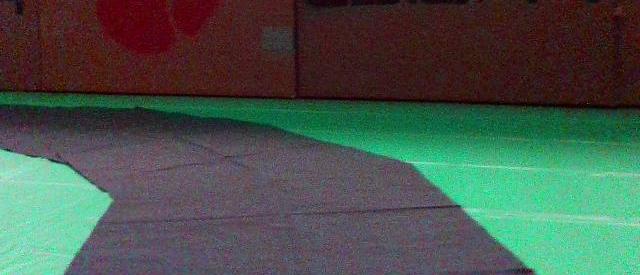}
         \caption{}
         \label{fig6b}
     \end{subfigure}
     \hfill
     \begin{subfigure}[b]{0.32\linewidth}
         \centering
         \includegraphics[width=\linewidth]{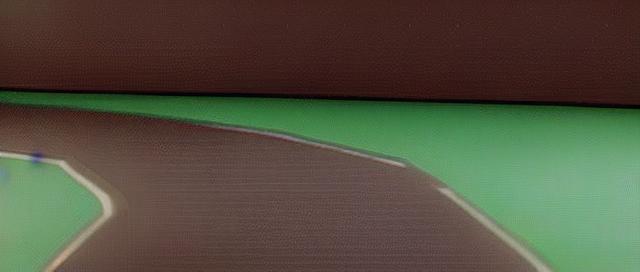}
         \caption{}
         \label{fig6c}
     \end{subfigure}

     \begin{subfigure}[b]{0.32\linewidth}
         \centering
         \includegraphics[width=\linewidth]{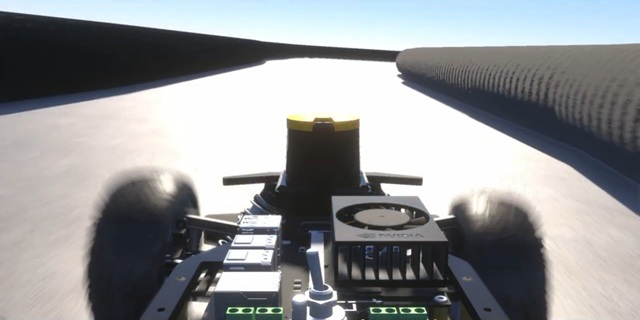}
         \caption{}
         \label{fig6d}
     \end{subfigure}
     \hfill
     \begin{subfigure}[b]{0.32\linewidth}
         \centering
         \includegraphics[width=\linewidth]{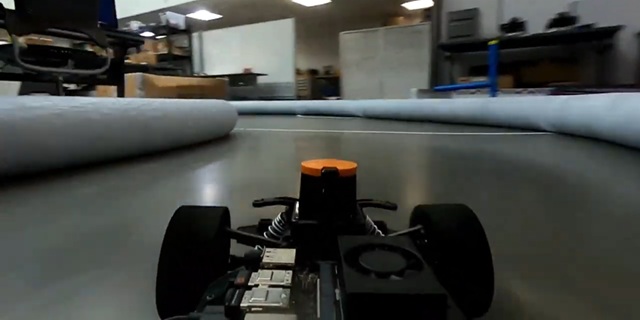}
         \caption{}
         \label{fig6e}
     \end{subfigure}
     \hfill
     \begin{subfigure}[b]{0.32\linewidth}
         \centering
         \includegraphics[width=\linewidth]{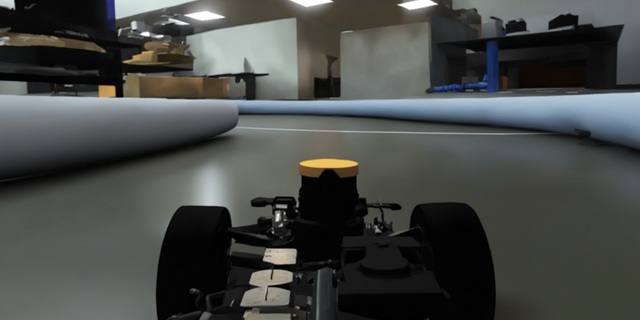}
         \caption{}
         \label{fig6f}
     \end{subfigure}

     \begin{subfigure}[b]{0.32\linewidth}
         \centering
         \includegraphics[width=\linewidth]{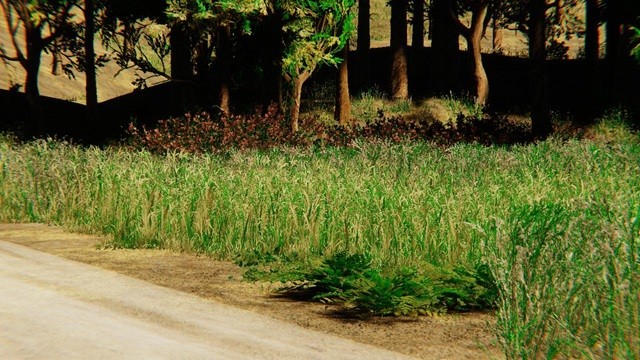}
         \caption{}
         \label{fig6g}
     \end{subfigure}
     \hfill
     \begin{subfigure}[b]{0.32\linewidth}
         \centering
         \includegraphics[width=\linewidth]{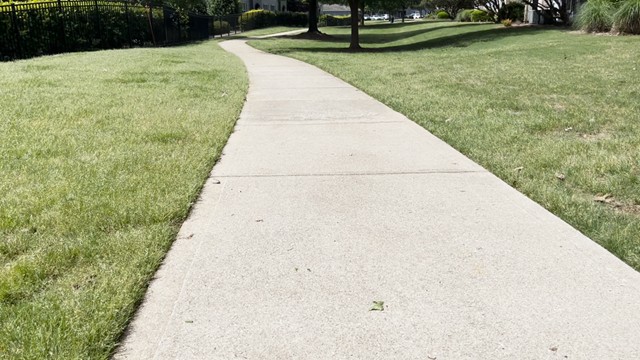}
         \caption{}
         \label{fig6h}
     \end{subfigure}
     \hfill
     \begin{subfigure}[b]{0.32\linewidth}
         \centering
         \includegraphics[width=\linewidth]{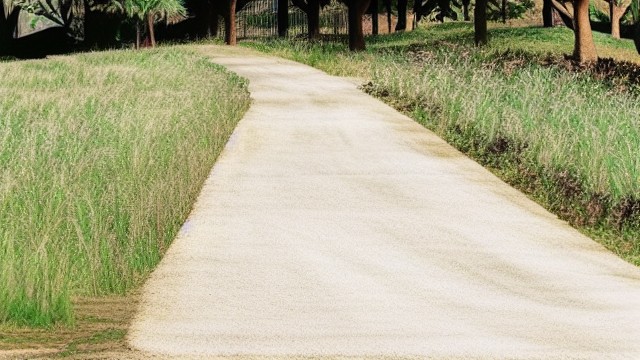}
         \caption{}
         \label{fig6i}
     \end{subfigure}
     \caption{Results of the 4$^\textrm{th}$ ablation study performed on fine-tuned model with text and image conditioning to qualitatively assess the domain adaptation capabilities: \texttt{<autodrive\_small\_onroad>} (a) prompt, (b) input, (c) output; \texttt{<autodrive\_small\_racing>} (d) prompt, (e) input, (f) output; \texttt{<autodrive\_large\_offroad>} (g) prompt, (h) input, (i) output.}
     \label{fig6}
\end{figure}

\begin{table}[t]
\centering
\caption{\small Quantitative Results of Ablation Study 4}
\label{tab5}
\resizebox{\columnwidth}{!}{%
\begin{tabular}{l|l|l|l|l|l}
    \hline
    \textbf{Domain}  &  \textbf{CS $\uparrow$}  & \textbf{CLIP-DS $\uparrow$} & \textbf{LPIPS $\downarrow$}  &  \textbf{SD-IS}  &  \textbf{SD-OS $\downarrow$}   \\ \hline
    Small, On-Road       & 7.58e-01    & 6.10e-02    & 3.74e-01   & 1.24e-03    & 6.89e-04   \\
    Small, Racing        & 7.36e-01    & 1.13e-01    & 1.26e-01   & 3.61e-04    & 3.18e-04   \\
    Large, Off-Road      & 8.59e-01    & 1.67e-01    & 3.82e-01   & 3.09e-03    & 1.93e-03   \\ \hline
\end{tabular}
}
\end{table}

\begin{figure*}[t]
     \centering
     \begin{subfigure}[b]{0.18\linewidth}
         \centering
         \includegraphics[width=\linewidth]{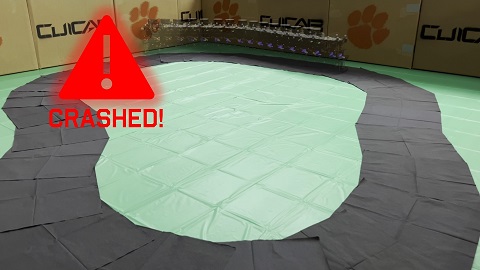}
         \caption{}
         \label{fig7a}
     \end{subfigure}
     \hfill
     \begin{subfigure}[b]{0.18\linewidth}
         \centering
         \includegraphics[width=\linewidth]{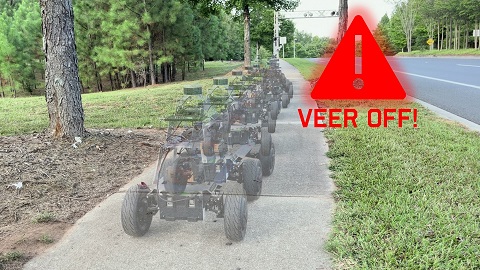}
         \caption{}
         \label{fig7b}
     \end{subfigure}
     \hfill
     \begin{subfigure}[b]{0.18\linewidth}
         \centering
         \includegraphics[width=\linewidth]{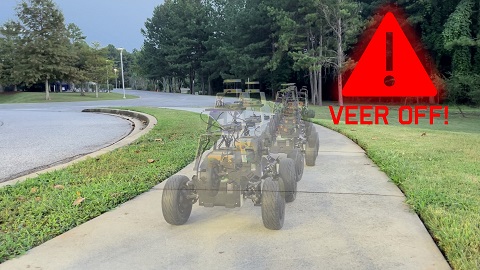}
         \caption{}
         \label{fig7c}
     \end{subfigure}
     \hfill
     \begin{subfigure}[b]{0.18\linewidth}
         \centering
         \includegraphics[width=\linewidth]{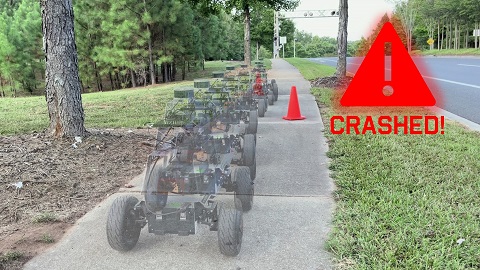}
         \caption{}
         \label{fig7d}
     \end{subfigure}
     \hfill
     \begin{subfigure}[b]{0.18\linewidth}
         \centering
         \includegraphics[width=\linewidth]{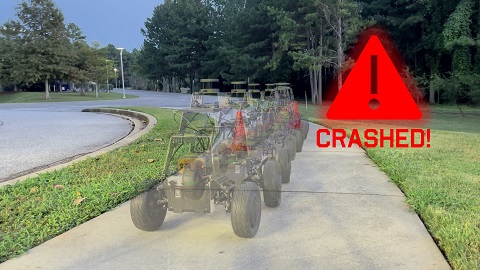}
         \caption{}
         \label{fig7e}
     \end{subfigure}
     \hfill
     \begin{subfigure}[b]{0.17\linewidth}
         \centering
         \includegraphics[width=\linewidth]{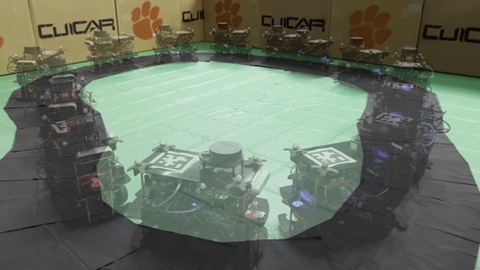}
         \caption{}
         \label{fig7f}
     \end{subfigure}
     \hfill
     \begin{subfigure}[b]{0.18\linewidth}
         \centering
         \includegraphics[width=\linewidth]{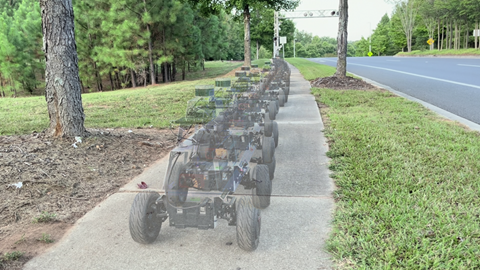}
         \caption{}
         \label{fig7g}
     \end{subfigure}
     \hfill
     \begin{subfigure}[b]{0.18\linewidth}
         \centering
         \includegraphics[width=\linewidth]{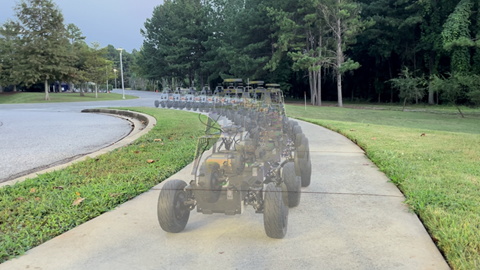}
         \caption{}
         \label{fig7h}
     \end{subfigure}
     \hfill
     \begin{subfigure}[b]{0.18\linewidth}
         \centering
         \includegraphics[width=\linewidth]{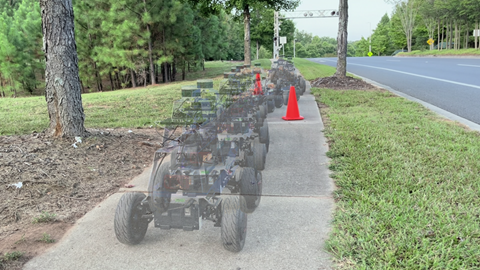}
         \caption{}
         \label{fig7i}
     \end{subfigure}
     \hfill
     \begin{subfigure}[b]{0.18\linewidth}
         \centering
         \includegraphics[width=\linewidth]{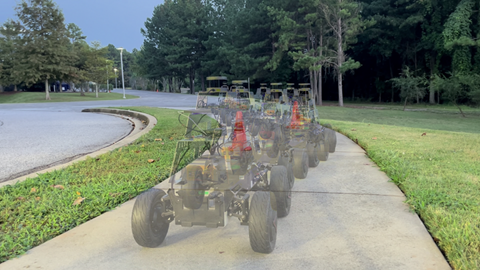}
         \caption{}
         \label{fig7j}
     \end{subfigure}
     \caption{Results of sim2real transfer of the end-to-end driving algorithms trained using simulation-only data without any augmentations: (a-e) exemplar instances where the vehicle fails to traverse the path when sim2real diffusion is OFF, and (f-j) successful autonomous driving in the real world when sim2real diffusion is ON. Nigel was deployed indoors while Hunter was deployed outdoors on straight (S) and curved (C) paths, with obstacles (O) in a subset of experiments.}
     \label{fig7}
\end{figure*}

The fourth ablation study (refer Fig. \ref{fig6} and Table \ref{tab5}) analyzed the effect of image prompt on the fine-tuned model in an image-to-image translation task. The focus of this study was to assess the semantic and geometric consistency during domain adaptation, conditioned by both text and image prompts. It is to be noted that the prompt image was randomly sampled from the pool of a few simulation images and had no paired relationship with the input image (which was sampled from the real-world data). Results of this ablation study (refer Fig. \ref{fig6}) show that the model preserves semantic and geometric features during domain adaptation, while ensuring adequate style transfer. Some of the prominent highlights from the on-road domain adaptation (Fig. \hyperref[fig6]{\ref*{fig6}(a-c)}) include generation of road lane markings, smoothing of the creases/reflections on the green table cloth and black tissues, removing box boundaries, toning the color of boxes closer to whiter shades, and removal of any graphical imprints on the boxes. For the racing domain adaptation (Fig. \hyperref[fig6]{\ref*{fig6}(d-f)}), some of the noticeable adaptations include smoothing ground and duct textures, preserving original vehicle and environment features, attenuating reflections/glare, and color tone-mapping (e.g., changing the color of LIDAR cap from orange to yellow). Finally, for the off-road domain adaptation (Fig. \hyperref[fig6]{\ref*{fig6}(g-i)}), we can notice that the model has preserved the path geometry and semantics, mapped the sidewalk pavement to a dirt road texture, adapted the mowed lawn to tall dry grass, and replaced the real trees with simulated ones.

\subsection{Performance Study}

The performance study (refer Table \ref{tab6}) captures training time and inference speed across 7 different compute settings. We can observe that fine-tuning the foundation models using the proposed pipeline is feasible across all computing platforms. The worst-case training time reported is just over 20 minutes (T4 GPU via Google Colab Free Tier), which ensures that the framework is serviceable to the community, who may not have access to high-performance computing resources. Additionally, compared to the training time for the behavioral cloning ($\sim$1.5 hours), fine-tuning the diffusion model is relatively trivial. Finally, the fine-tuning process requires only a handful (5-10) domain-specific examples from simulation to learn the new concept. This ensures that the proposed pipeline is not only fast and convenient but also highly scalable across different tasks and domains.

It is also worth noting that the model can provide pseudo-real-time inference even on local (3080 Ti) and edge (2060) compute resources. As highlighted earlier, slow-speed autonomous systems (e.g., Nigel/Hunter in our case) can run sim2real diffusion to complete their objective in real-time. Distributed computing frameworks can alleviate any edge-computing limitations (memory, throughput, etc.), potentially supporting faster-than-real-time inference (e.g., $>$65 IPS with L40S nodes of Palmetto Cluster).

\begin{table}[t]
\centering
\caption{\small Performance Evaluation of Sim2Real Diffusion}
\label{tab6}
\resizebox{\columnwidth}{!}{%
\begin{tabular}{l|l|l|l|l|l|l|l}
    \hline
    \textbf{Evaluation Phase}  &  \textbf{2060} &  \textbf{3080 Ti}  & \textbf{T4}  & \textbf{V100}  & \textbf{A100}  & \textbf{H100} & \textbf{L40S} \\ \hline
    Training (mm:ss) $\downarrow$  & 17:26  & 13:33  & 21:42  & 05:12  & 02:58  & 02:37  & 02:26 \\
    Inference (IPS) $\uparrow$   & 05.83  & 07.07  & 03.76  & 23.43  & 32.52  & 49.69  & 65.07 \\ \hline
\end{tabular}
}
\end{table}

\begin{table}[t]
\centering
\caption{\small Quantitative Evaluation of Sim2Real Diffusion}
\label{tab7}
\resizebox{\columnwidth}{!}{%
\begin{tabular}{l|l|l|l|l|l|l|l|l}
    \hline
    \textbf{Deployment}  &  \textbf{FID $\downarrow$}  &  \textbf{KID $\downarrow$}  &  \textbf{CS $\uparrow$}  & \textbf{CLIP-DS $\uparrow$} & \textbf{LPIPS $\downarrow$}  & \textbf{SD-IS}  & \textbf{SD-OS $\downarrow$}    & \textbf{IPS$^\mathrm{\dagger}$ $\uparrow$}   \\ \hline
    Nigel (Indoor)         & 12.02 & 6.48e-03 & 8.15e-01 & 1.08e-01 & 4.06e-01  & 1.32e-03  & 7.89e-04    & 7.07    \\ 
    Hunter (Outdoor-S)     & 13.16 & 8.97e-03 & 7.31e-01 & 1.23e-01 & 3.14e-01  & 1.83e-03  & 1.07e-03    & 5.41    \\ 
    Hunter (Outdoor-C)     & 11.89 & 9.36e-03 & 7.22e-01 & 1.29e-01 & 3.97e-01  & 1.75e-03  & 9.84e-04    & 6.09    \\ 
    Hunter (Outdoor-SO)    & 14.26 & 1.58e-02 & 7.57e-01 & 1.18e-01 & 3.05e-01  & 8.94e-03  & 5.12e-03    & 6.11    \\ 
    Hunter (Outdoor-CO)    & 15.69 & 1.76e-02 & 7.46e-01 & 1.16e-01 & 3.10e-01  & 8.62e-03  & 4.93e-03    & 5.69    \\ \hline  
    \multicolumn{9}{l}{\footnotesize $^\mathrm{\dagger}$Nigel: 3080 Ti (distributed compute), Hunter: 2060 (on-board edge compute).}
\end{tabular}
}
\end{table}

\subsection{Case Studies}

The exemplar case studies serve the purpose of demonstrating the serviceability of the proposed sim2real transfer method (refer Fig. \ref{fig7}, Table \ref{tab7}, and Table \ref{tab8}). It is worth mentioning that while an earlier work \cite{AutoDRIVE2023} has demonstrated zero-shot transferability of behavioral cloning, we deliberately exaggerated the sim2real gap by choosing a dissimilar real-world setup and pruning the data augmentation step. This allowed us to demonstrate the efficacy of sim2real diffusion during deployment, while also reducing the data requirement for behavioral cloning by $\sim64\times$ ($>50\times$ reduction in training time) without compromising on the sim2real transfer.

When sim2real diffusion is turned off, the model frequently fails to navigate the designated path properly, where the vehicle veers off the course, potentially leading to collisions. These failures are primarily attributed to the distributional shift between the synthetic training domain and the complexities of the real-world domain (i.e., sim2real gap). Without adaptation, the model overfits to simulation-specific artifacts and fails to learn representations that are robust to real-world variations such as changes in lighting, textures, colors, reflections, glare, sensor noise, etc.

In contrast, the same behavioral cloning model, when deployed with sim2real diffusion enabled, is capable of completing multiple autonomous runs in indoor/outdoor real-world experimental setups with small/large vehicles. This improvement can be attributed to the diffusion process aligning the synthetic and real data distributions through learned image-level transformations and feature-space adaptations, thereby reducing the domain discrepancy at both the input and intermediate representation levels. These empirical results support the hypothesis that sim2real diffusion acts as an effective domain adaptation strategy (40.33\% reduction in mean style difference, i.e., $\frac{\mu_{\mathrm{SD\textrm{-}IS}}-\mu_{\mathrm{SD\textrm{-}OS}}}{\mu_{\mathrm{SD\textrm{-}IS}}}$), allowing reliable policy transfer from simulation to reality.

Finally, from comparison with the baselines (refer Table \ref{tab8}), we can see that domain randomization may work but possesses no transferability guarantees. While CycleGANs remain practical for faster inference times, the merits of the proposed sim2real diffusion far outweigh this benefit, making it a stronger candidate.

\begin{table}[t]
\centering
\caption{\small Benchmarking Sim2Real Diffusion with Baselines}
\label{tab8}
\resizebox{\columnwidth}{!}{%
\begin{tabular}{l|l|l|l|l|l|l}
    \hline
    \textbf{Approach}  &  \textbf{Data $\downarrow$}  & \textbf{TT$^\mathrm{\dagger}$ $\downarrow$} & \textbf{IPS$^\mathrm{\ddagger}$ $\uparrow$}  &  \textbf{PAT$^\mathrm{\star}$ $\uparrow$}  &  \textbf{RMS-XTE$^\mathrm{\star}$ $\downarrow$}  &  \textbf{SR$^\mathrm{\star}$ $\uparrow$}   \\ \hline
    \multicolumn{7}{l}{\textbf{Nigel (Indoor)}} \\ \hline
    Domain Randomization         & 108k        & 4.8 hrs    & 26.16   & 78.24\%    & 5.09e-02    & 80\%   \\
    CycleGAN                     & 3.4k        & 7.6 hrs    & 18.69   & 89.32\%    & 3.52e-02    & 90\%   \\
    Sim2Real Diffusion (Ours)    & 10          & 0.2 hrs    & 07.07   & 95.61\%    & 2.43e-02    & 90\%   \\ \hline
    \multicolumn{7}{l}{\textbf{Hunter (Outdoor-S)}} \\ \hline
    Domain Randomization         & 192k        & 8.6 hrs    & 21.56   & 72.47\%    & 2.72e-01    & 60\%   \\
    CycleGAN                     & 6.0k        & 14.1 hrs   & 14.64   & 82.26\%    & 1.86e-01    & 70\%   \\
    Sim2Real Diffusion (Ours)    & 10          & 0.2 hrs    & 05.83   & 90.15\%    & 8.37e-02    & 90\%   \\ \hline
    \multicolumn{7}{l}{\footnotesize $^\mathrm{\dagger}$Training time for base behavioral cloning model was 1.2 hrs for Nigel and 1.9 hrs for Hunter with 3080 Ti.} \\
    \multicolumn{7}{l}{\footnotesize $^\mathrm{\ddagger}$Nigel: 3080 Ti (distributed compute), Hunter: 2060 (on-board edge compute).} \\
    \multicolumn{7}{l}{\footnotesize $^\mathrm{\star}$Results aggregated over 10 trials.}    
\end{tabular}
}
\end{table}


\section{Conclusion}
\label{Section: Conclusion}

In this work, we addressed some of the limitations of existing sim2real transfer approaches for autonomous driving by proposing a unified framework based on conditional latent diffusion models. Our framework specifically targets autonomy-oriented requirements — namely, the need for conditioned domain adaptation, few-shot generalization, and modular handling of multiple domain shifts — while remaining scalable and real-time executable. Through extensive ablation studies, performance evaluations, and experiments, we demonstrated the efficacy of our approach in bridging the perceptual domain gap between simulated and real-world data distributions by over 40\%. These findings highlight the potential of leveraging diffusion models as a flexible and scalable approach to sim2real transfer.

Since high-speed applications of the current framework may need better compute resources for real-time performance, future research could attempt to further improve the real-time performance of the proposed pipeline via time or knowledge distillation techniques. Additionally, we wish to expand the existing framework to learn more simulation styles across different vehicles, environments, and ODDs. Finally, incorporating custom guardrails and physics-based machine learning principles can potentially improve the domain adaptation capabilities with trustworthy grounding.


\balance
\bibliographystyle{IEEEtran}
\bibliography{references}

\end{document}